\documentclass[11pt]{article}
\usepackage{amsmath}
\usepackage[margin=1in]{geometry}
\usepackage{booktabs}
\usepackage{multirow}
\usepackage{graphicx}
\usepackage[hidelinks]{hyperref}
\usepackage{amssymb}

\newcommand{\approach}{\textit{Stepwise-Querying}}

\usepackage[preprint]{acl}

\usepackage{times}
\usepackage{latexsym}

\usepackage[T1]{fontenc}

\usepackage[utf8]{inputenc}

\usepackage{microtype}

\usepackage{inconsolata}

\usepackage{graphicx}

\title{\approach{}: Distilling Reasoning without Access to Teacher Chain-of-Thought}

\author{Dicong Yu, Wenhua Wang, Tian Wang \\
  Institute of AI and Future Network, Beijing Normal University
}

\begin{document}
\maketitle

\begin{abstract}
Reasoning distillation is widely used to transfer the reasoning capabilities of large language models to smaller students through teacher-generated supervision. In the black-box setting, however, teachers may expose only final answers and brief solutions, leaving intermediate calculations and validity conditions unavailable and thus limiting the reasoning supervision available for distillation. We propose \approach{}, which elicits step-specific explanations without requiring access to the teacher's internal chain-of-thought (CoT) traces. Independent queries to the same teacher request intermediate derivations, validity checks, and corrections for individual steps, the assembled solutions are then used for standard supervised fine-tuning. Across two teachers, two students, and four mathematics and code benchmarks, our method achieves the highest score in 14 of 16 comparisons with three baselines. We further introduce Stepwise Explanation Information Gain (SEIG), a diagnostic based on the log-ratio of a frozen student's probabilities of correct step completion under stepwise and global explanations. Analyses on GSM8K show that incorrect-answer trajectories can remain useful when they preserve valid intermediate reasoning, while further decomposition yields limited additional gains.
\end{abstract}

\section{Introduction}

Reasoning distillation transfers capabilities to smaller language models through teacher-generated reasoning supervision \cite{magister2023teaching}. In a black-box settings, however, teachers may expose only final answers and brief solutions, leaving intermediate calculations and validity conditions unavailable \cite{zhang2026steal}, which limits the reasoning supervision available for distillation. MathFimer expands mathematical solutions with a trained fill-in-the-middle model \cite{yan2026mathfimer}, while Trace Inversion synthesizes detailed traces from observable questions, answers, and optional summaries \cite{zhang2026steal}. These approaches motivate studying how further queries to the same teacher can turn brief solutions into useful training data.

The quality of such data depends on both query organization and prompting. Research on reverse chain-of-thought generation identifies post-hoc rationalization \cite{peng2026rationalization}, and answer-conditioned distillation experiments distinguish instructions to rationalize toward an answer from answer visibility alone \cite{lee2026answerconditioned}. A simple example illustrates the problem: for $\sqrt{(x-1)^2}=2$, an initial answer \(x=3\) is valid but incomplete. A request to explain this answer may produce a correct substitution without discovering \(x=-1\). A prompt requesting independent derivation, enumeration of all cases, and correction of the candidate instead targets the complete solution. This example motivates a comparison; it does not establish that either prompt reliably produces a particular outcome. 
\begin{figure*}[t]
    \centering
    \includegraphics[width=\textwidth]{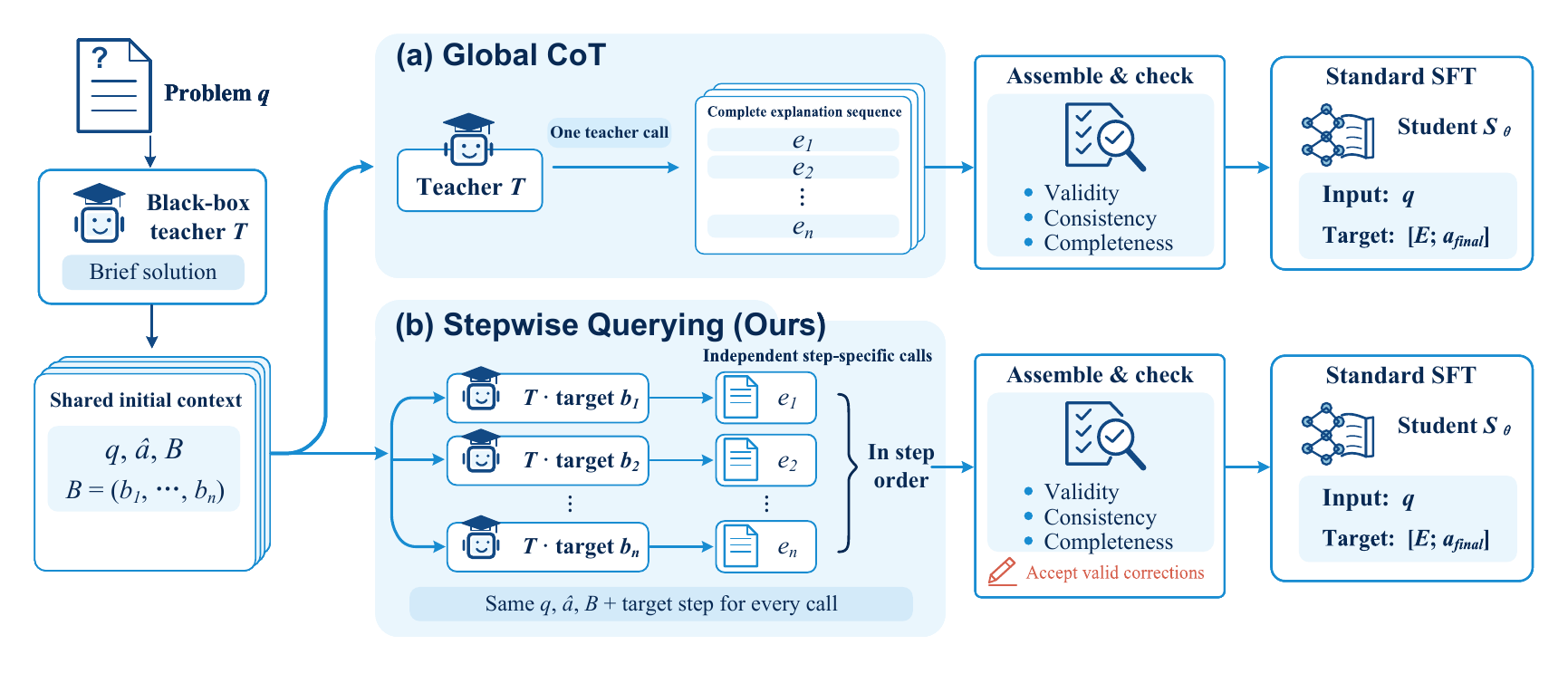}
    \caption{Comparison of supervision construction. (a) Global CoT: a single teacher call produces one explanation for all steps. (b) Stepwise-Querying (ours): independent calls, each targeting one step under a shared initial context, are assembled, checked for validity, consistency, and completeness, and used for standard SFT.}
    \label{fig:1}
\end{figure*}

Considering the above observation, we propose \approach{}, which targets independent derivation and correction with prompts. Firstly, the teacher provides a candidate answer and brief solution steps. Then, subsequent calls target individual steps, requesting intermediate calculations, mathematical justifications, and necessary conditions, and each call uses an independent context containing the original problem and initial solution. 
The prompts treat the initial candidate answer and steps as provisional, potentially incorrect or incomplete. They ask the teacher to derive the target step from the problem before comparing it with the candidate, and they allow the teacher to identify missing cases and propose corrections.
The expanded solutions are then checked for consistency and completeness before being used for standard supervised fine-tuning, where the student receives only the original problem as input. 
Our study asks whether combining multiple targeted queries with these prompts improves supervision quality and student reasoning under controlled costs (Figure~\ref{fig:1}). 



Our methodological contributions are:
\begin{itemize}
    \item \textbf{A stepwise querying protocol for black-box reasoning distillation.} We turn brief teacher solutions into detailed training data without access to internal CoT traces: each step receives an independent query whose prompt asks the teacher to derive the step from the problem, verify its validity and completeness, and correct the provisional candidate answer—rather than to rationalize toward it. The protocol applies to any multi-step task, from mathematics to code generation.


    \item \textbf{Stepwise Explanation Information Gain (SEIG), a diagnostic for local supervisory value.} SEIG measures the log-ratio of a frozen student's probability of correct step completion under stepwise versus global CoT explanations, using only public explanation text and independent verification labels, with no teacher logits or hidden traces required.

    
    \item \textbf{A controlled evaluation with actionable findings.} Across two teachers, two students, and four benchmarks, our method achieves the best score in 14 of 16 comparisons with three published baselines and outperforms global CoT and answer-explanation supervision in all 16 settings. SEIG analyses further show that incorrect-answer trajectories remain useful when they preserve valid intermediate reasoning, and that recursive decomposition beyond the step level yields diminishing returns.
    
\end{itemize}

\section{Related Work}

Reasoning distillation and step expansion. Magister et al. transfer reasoning capabilities to smaller models by training on teacher-generated chains of thought \cite{magister2023teaching}. MathFimer \cite{yan2026mathfimer} expands mathematical solutions by training a fill-in-the-middle model to insert intermediate steps. Zhang et al. synthesize reasoning traces from problems, answers, and optional summaries using a trained trace inversion model \cite{zhang2026steal}. These studies motivate constructing detailed reasoning supervision from available outputs. Our proposed method focuses on the querying protocol: it uses multiple teacher calls to expand individual steps and combines the resulting derivations into student training examples.

Decomposed querying and iterative refinement. Least-to-Most Prompting \cite{zhou2022least} decomposes a problem into subproblems and solves them sequentially using previous answers. Self-Refine \cite{madaan2023selfrefine} iteratively generates feedback and revises an output with the same language model. Our protocol instead assigns a specific mathematical step to each follow-up call, while retaining the same original problem and initial solution context. Newly expanded steps are not passed to subsequent calls. This distinguishes independent step expansion for distillation data construction from sequential subproblem solving and feedback-driven revision.

Answer-conditioned generation and rationalization. STaR \cite{zelikman2022star} regenerates unsuccessful rationales with the correct answer as a hint and trains on successful generations. Peng et al.  study post-hoc rationalization and introduce answer-invariant structural skeletons to guide generation \cite{peng2026rationalization}. Lee et al. show that answer-conditioned rationales can impair distillation despite final-answer correctness filtering, identifying instructions to rationalize toward an answer as a major source of harm \cite{lee2026answerconditioned}. Motivated by these findings, our prompts treat initial outputs as provisional and request derivation from the problem, validity checks, complete case coverage, and correction. Whether these instructions improve reasoning supervision and student performance remains an empirical question.

\section{Method} \label{sec:method}

\subsection{Problem Formulation and Overview} \label{sec:problem}

We study reasoning distillation for problems that admit multi-step solutions, using a black-box teacher \(T\) and a student model \(S_\theta\), where \(\theta\) denotes the student's trainable parameters. The formulation applies to tasks such as mathematical reasoning, logical inference, commonsense reasoning, and multi-hop
question answering. Access to the teacher is limited to its public text-generation interface. Given a problem \(q\), the teacher generates a candidate answer \(\hat{a}\) and an ordered list of brief reasoning steps \(B=(b_1,\ldots,b_n)\). Each step represents a coherent reasoning operation. The initial generation is expressed as
\begin{equation}
    (\hat{a},B)=T(q;\pi_{\mathrm{brief}}),
    \label{eq:brief_generation}
\end{equation}
where \(\pi_{\mathrm{brief}}\) requests a concise solution.

Our framework consists of four stages: \textbf{brief solution generation, stepwise querying, assembly and
checking, and student supervised fine-tuning}. For each step \(b_i\), a separate teacher call generates an expanded derivation:
\begin{equation}
    e_i=T(q,\hat{a},B,i;\pi_{\mathrm{expand}}),
    \qquad i=1,\ldots,n,
    \label{eq:stepwise_expansion}
\end{equation}
where \(\pi_{\mathrm{expand}}\) requests intermediate inferences, supporting justifications, and relevant conditions. The candidate answer and initial steps serve as provisional references. Each query uses the same initial context, without receiving expansions generated by other calls.

The expansions are assembled in order into a detailed solution \(E\) and screened using task-appropriate criteria for step validity, consistency, and completeness. Retained training examples take the form \((q,y)\), where \(y=[E;a_{\mathrm{final}}]\) concatenates the expanded reasoning and final answer. Here, \(a_{\mathrm{final}}\) denotes the answer determined during assembly and checking, which may differ from the initial candidate answer \(\hat{a}\). The student learns to generate \(y\) from the original problem \(q\) alone. This framework constructs detailed reasoning supervision from the teacher's publicly available outputs.

\subsection{Stepwise Querying} \label{sec:stepwise-quering}

\subsubsection{Initial Answer and Brief Step Generation}
\label{sec:initial_generation}

Using \(\pi_{\mathrm{brief}}\), we first query the teacher \(T\) with the original problem \(q\) to obtain a candidate answer \(\hat{a}\) and numbered reasoning steps \(B=(b_1,\ldots,b_n)\). The prompt requests a complete answer while keeping the reasoning concise. Each step should correspond to a substantive reasoning operation, such as drawing an inference, integrating relevant information, or considering alternative cases. Detailed derivations and justifications are reserved for subsequent stepwise queries.

We retain candidate answers even when they are incorrect or incomplete. Initial processing checks only whether the response contains an identifiable candidate answer and a non-empty, parseable sequence of reasoning steps; answer correctness is not an inclusion criterion. Both the candidate answer and the initial steps are treated as provisional information that may contain errors or omissions. Subsequent queries are allowed to question these outputs and propose corrections.

Each structurally usable initial output is stored and reused unchanged as the common reference for all follow-up queries. Generation variants compared in our experiments also share the same initial output, allowing differences in querying protocols and prompts to be examined from a common starting point.

\subsubsection{Prompt Design for Derivation and Correction}

\begin{figure*}[t]
    \centering
    \includegraphics[width=\textwidth]{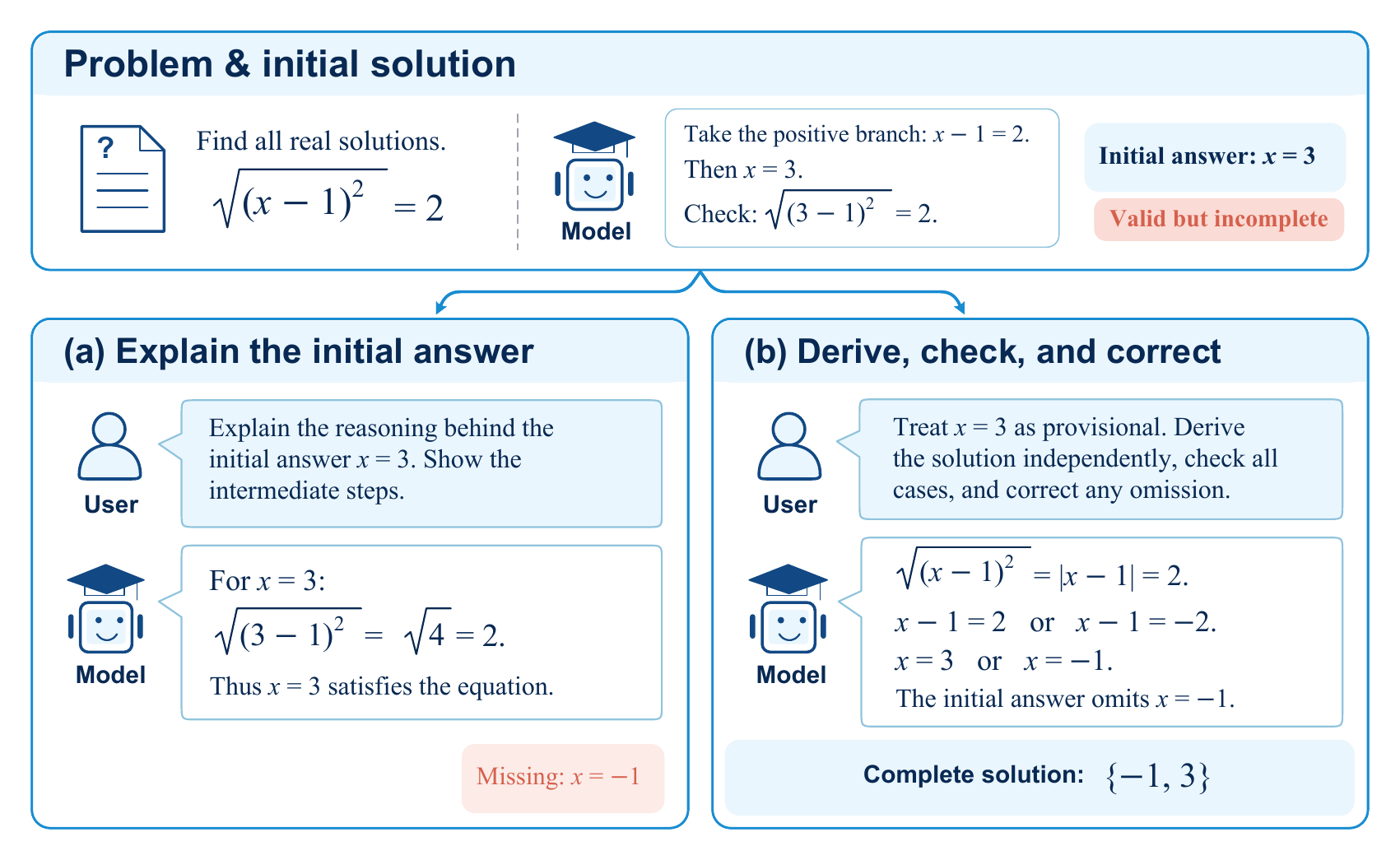}
    \caption{Illustrative comparison of prompting strategies.
    Explaining the initial answer verifies \(x=3\) but omits
    \(x=-1\), while independent derivation and completeness
    checks recover both roots.}
    \label{fig:derivation_correction}
\end{figure*}

We design \(\pi_{\mathrm{expand}}\) to treat the initial answer and brief steps as provisional references. The prompt asks the teacher to derive the target step from the problem and relevant premises, then compare the resulting conclusion with the initial solution. The candidate answer should be used for checking the
result rather than as a premise that establishes its validity. As illustrated in Figure~\ref{fig:derivation_correction}, explaining the initial answer can overlook a missing solution, while independent derivation and completeness checks recover both roots.

The prompt also requests task-specific validity and completeness checks. For mathematical tasks, these include domain constraints, sign restrictions, conditions for equivalent transformations, and the applicability of theorems. For logical or evidence-based tasks, they include premise consistency and support from the available information. Steps involving case analysis or answer aggregation must cover all cases or components required by the problem.

When an error or omission is detected, the teacher is instructed to identify the affected claim, explain the issue, and propose a correction. Proposed corrections are recorded in the response for subsequent assembly and checking. The prompt additionally requires consistent notation and terminology, a focus on the target step, and minimal repetition of other steps.

\subsubsection{Student Model Training}

For each problem \(q\), we organize the expanded responses \(e_1,\ldots,e_n\) according to their original step indices. Explicit correction proposals are incorporated into the affected claims after checking their validity and consistency with the remaining steps. The final answer \(a_{\mathrm{final}}\) is taken from an accepted answer correction when available; otherwise, the initial candidate \(\hat{a}\) is retained if it passes the final checks. Samples with unresolved contradictions or conflicting answer corrections are excluded from the training set. This screening is performed after expansion, allowing incorrect initial answers to contribute training examples through correction.

We normalize the retained responses by removing query metadata, duplicate headings, and redundant final-answer statements. Let \(\tilde{e}_i\) denote the normalized response for step \(i\).
The detailed solution is constructed as \(E=\mathrm{Join}(\tilde{e}_1,\ldots,\tilde{e}_n)\), where
\(\mathrm{Join}\) concatenates the steps in order with consistent formatting. We append the final answer once at the end:
\begin{equation}
    y=[E;a_{\mathrm{final}}].
    \label{eq:training_target}
\end{equation}
Each training example contains the original problem \(q\) as input and \(y\) as the supervised output. The initial answer, brief steps, and follow-up dialogue serve as generation context
and are not separately included in the student's input.

Given the resulting dataset \(\mathcal{D}\), we train \(S_\theta\) using the standard supervised fine-tuning objective:
\begin{equation}
    \mathcal{L}_{\mathrm{SFT}}(\theta)
    =
    -\frac{1}{|\mathcal{D}|}
    \sum_{(q,y)\in\mathcal{D}}
    \sum_{t=1}^{|y|}
    \log p_\theta(y_t\mid q,y_{<t}),
    \label{eq:sft_loss}
\end{equation}
where \(y_t\) is the \(t\)-th output token. The loss is applied only to output tokens; input and padding tokens are masked.

Sequence lengths are measured using the student's tokenizer and, when applicable, its chat template. Only examples whose complete input-output sequences fit the training context limit are used for fine-tuning. Overlength examples remain in the generated data archive but are excluded from the training set,
preserving complete reasoning and the final answer.

\subsection{Stepwise Explanation Information Gain} \label{sec:seig}

While Section \ref{sec:stepwise-quering} specify how to construct detailed supervision from a black-box teacher and train the student, they do not characterize how helpful different teacher explanations are at the level of individual reasoning steps. We therefore introduce Stepwise Explanation Information Gain (SEIG), a diagnostic that measures the local supervisory value of stepwise explanations relative to globally generated chain-of-thought (CoT) explanations. SEIG is not a training objective and does not alter the data-construction or supervised fine-tuning procedures above. Instead, it provides an evaluation protocol that compares two explanations of the same step under a frozen student and shared local conditions. This section defines the shared step setup, the step-level SEIG, and its empirical estimation and aggregation. All comparisons use only the teacher's public explanation text, student completions, and independent verification labels, without requiring teacher logits or hidden CoT traces.


\subsubsection{Shared steps and explanation generation.}
Both methods explain the same ordered step list:
\begin{equation}
B=(b_1,b_2,\ldots,b_n).
\end{equation}
For each sample $k=1,\ldots,K$, the explanation sequences are
\begin{equation}
\begin{aligned}
E_k^{\mathrm{CoT}}
&=(e_{1,k}^{\mathrm{CoT}},\ldots,e_{n,k}^{\mathrm{CoT}}),\\
E_k^{\mathrm{Step}}
&=(e_{1,k}^{\mathrm{Step}},\ldots,e_{n,k}^{\mathrm{Step}}).
\end{aligned}
\end{equation}
Both $e_{i,k}^{\mathrm{CoT}}$ and $e_{i,k}^{\mathrm{Step}}$ explain the same step $b_i$. CoT generates each complete sequence in one teacher call, whereas Step generates its component explanations through independent step-specific calls. Each method therefore provides $K$ teacher-generated explanations per shared step, compared using the common step index $i$.

\subsubsection{Common local conditions and SEIG.}
Both conditions use the same local context $c_i$, correct target $z_i^\ast$, and frozen student $R_\phi$. The local context contains the problem, shared verified preceding conclusions, and the current step task:
\begin{equation}
c_i=[q;r_{<i};h_i].
\end{equation}
Applying the same target-masking rule produces $\tilde e_{i,k}^{\mathrm{CoT}}$ and
$\tilde e_{i,k}^{\mathrm{Step}}$. Each explanation is evaluated with one student completion using the common context $c_i$.

Let $\mathcal G_i^m$ denote the distribution of masked teacher explanations for step $b_i$ induced by method
$m\in\{\mathrm{CoT},\mathrm{Step}\}$ under the shared initial context. Under the fixed student decoding protocol, the success probability is
\begin{equation}
P_i^m = \mathbb{E}_{\tilde e\sim\mathcal G_i^m} \left[ P_{R_\phi}(z_i^\ast\mid c_i,\tilde e) \right].
\end{equation}
The inner probability refers to successful completion under the shared acceptance criterion, allowing equivalent correct expressions or implementations. Thus, $P_i^m$ averages the student's success probability over the teacher-explanation distribution for that method.
The step-level SEIG is
\begin{equation}
\operatorname{SEIG}_i
=
\log_2\frac{P_i^{\mathrm{Step}}}{P_i^{\mathrm{CoT}}}.
\end{equation}
A positive value indicates that Step explanations are more helpful for correct step completion than the corresponding CoT explanations. For empirical measurement, each of the $K$ teacher trajectories is
evaluated with one student completion, and the two probabilities are estimated using the same success-counting and smoothing procedure.

\subsubsection{Aggregation.}
For problem $q$ with $n_q$ shared steps, we first average across
steps and then across the $N$ problems in the analysis set
$\mathcal Q$:
\begin{equation}
\begin{aligned}
\operatorname{SEIG}(q)
&=\frac{1}{n_q}
  \sum_{i=1}^{n_q}\operatorname{SEIG}_i,\\
\overline{\operatorname{SEIG}}
&=\frac{1}{N}
  \sum_{q\in\mathcal Q}\operatorname{SEIG}(q).
\end{aligned}
\end{equation}
This assigns equal weight to each problem. The same aggregation is applied to empirical SEIG estimates.

\subsubsection{Probability Estimation over Sampled Teacher Trajectories.}
For each shared step $b_i$, both CoT and Step sample $K$ teacher-generated reasoning trajectories. Let $e_{i,k}^m$ denote the explanation of step $b_i$ from the $k$-th trajectory under method $m\in\{\mathrm{CoT},\mathrm{Step}\}$. For CoT, this explanation is taken from the corresponding step within a globally generated trajectory; for Step, it is generated by a query targeting that step. We apply the same masking rule to remove direct statements of the target conclusion and answer leakage, obtaining $\tilde e_{i,k}^m$.

Each explanation is evaluated using the same frozen student $R_\phi$ and shared local context $c_i$, which contains the problem, verified preceding conclusions, and the current step task.
The student inputs are
\begin{equation}
\begin{aligned}
x_{i,k}^{\mathrm{CoT}} &= [c_i;\tilde e_{i,k}^{\mathrm{CoT}}],\\
x_{i,k}^{\mathrm{Step}} &= [c_i;\tilde e_{i,k}^{\mathrm{Step}}],
\end{aligned}
\label{eq:seig_sampling_inputs}
\end{equation}
where $k=1,\ldots,K$. The two conditions share the same step task and correct target $z_i^\ast$, which is not supplied to the student. We generate exactly one student completion for each
sampled teacher trajectory:
\begin{equation}
z_{i,k}^m\sim R_\phi(\cdot\mid x_{i,k}^m),
\qquad k=1,\ldots,K.
\label{eq:seig_sampling}
\end{equation}
Thus, $K$ counts teacher trajectories; the $K$ student completions result from evaluating those trajectories individually. The student checkpoint, output instructions, temperature, top-$p$, and maximum generation length are identical across the two conditions.

An independent verifier $V_i$ evaluates each completion against the same predefined requirements associated with $z_i^\ast$:
\begin{equation}
V_i(z)=
\begin{cases}
1, & \text{if $z$ satisfies the requirements of step $b_i$},\\
0, & \text{otherwise}.
\end{cases}
\label{eq:seig_verifier}
\end{equation}
Equivalent correct expressions or implementations are accepted. Outputs with unresolved correctness are reviewed before binary labels are assigned. The number of successful trajectory--completion
pairs is
\begin{equation}
k_i^m=\sum_{k=1}^{K}V_i(z_{i,k}^m).
\label{eq:seig_success_count}
\end{equation}

Here, $P_i^m$ denotes the probability of correct step completion under the teacher-explanation distribution induced by method $m$ and the fixed student decoding protocol. Its empirical estimate is $k_i^m/K$. To avoid undefined or infinite logarithmic ratios when no successful completions are observed, we use
\begin{equation}
\widehat P_i^m=\frac{k_i^m+0.5}{K+1}.
\label{eq:seig_estimate}
\end{equation}
The empirical step-level SEIG is then
\begin{equation}
\widehat{\operatorname{SEIG}}_i
=
\log_2
\frac{\widehat P_i^{\mathrm{Step}}}
     {\widehat P_i^{\mathrm{CoT}}}.
\label{eq:seig_empirical}
\end{equation}
A positive value indicates that stepwise explanations yield a higher probability of correct step completion than the corresponding CoT explanations under the shared evaluation conditions and equal numbers of sampled teacher trajectories. The protocol uses teacher explanation text, student completions, and independent verification labels, without requiring teacher logits.

\section{Experiments}

\subsection{Experimental Setup}

\subsubsection{Teacher Models}

\paragraph{DeepSeek-V4-Pro.}
DeepSeek-V4-Pro is a mixture-of-experts model in the DeepSeek-V4 family, designed for reasoning, coding, and efficient long-context processing \citep{deepseek2026deepseek}. We use it to generate initial brief solutions and subsequent solution expansions.

\paragraph{GLM-5.2.}
GLM-5.2 is a model designed for long-horizon tasks, with support for a one-million-token context and coding-oriented capabilities \citep{glm5team2026glm5vibecodingagentic}. Following the citation recommendation in its official model card, we cite the GLM-5 technical report for the model family \citep{glm5team2026glm5vibecodingagentic} and the GLM-5.2 model card for version-specific information.

Both teachers are accessed as black-box text generators: the distillation protocol uses their public solution text without requiring model weights, logits, or access to hidden reasoning. Each teacher is evaluated separately under the same generation protocols. Teacher endpoint versions, query dates, reasoning settings, and sampling configurations are recorded for reproducibility.

\subsubsection{Student Models}
\paragraph{Qwen2.5-7B-Instruct.}
Qwen2.5-7B-Instruct is the instruction-tuned 7B-scale member of the Qwen2.5 family, which supports general language understanding, mathematical reasoning, and coding \citep{qwen2025qwen25technicalreport}. We use its released
instruction-tuned checkpoint as one student initialization.

\paragraph{Llama-3.1-8B-Instruct.}
Llama-3.1-8B-Instruct is Meta's instruction-tuned 8B model, belonging to the multilingual Llama 3 family described in \citet{grattafiori2024llama}. It provides a second student family for examining whether the effects of stepwise querying transfer across model families.

Each student is fine-tuned with standard supervised learning, using only the original problem as input and the synthesized detailed solution and final supervision answer as the target. Within each student family, all data-generation conditions share the same initial checkpoint and training recipe.

\subsubsection{Evaluation Datasets}

We evaluate mathematical reasoning on GSM8K and MATH-500, and code generation on HumanEval and LiveCodeBench.

\paragraph{GSM8K.}
GSM8K contains linguistically diverse grade-school mathematics word problems requiring multiple arithmetic reasoning steps \cite{cobbe2021training}. We use its official test split to evaluate students' ability to solve mathematical word problems.

\paragraph{MATH-500.}
MATH-500 is a 500-problem evaluation subset of the competition-level MATH benchmark \cite{hendrycks2021measuring}. It follows the held-out subset used in \citet{lightman2024let} and evaluates mathematical problem solving beyond elementary arithmetic.

\paragraph{HumanEval.}
HumanEval contains 164 hand-written Python programming tasks that require completing a function from its signature and docstring \citep{chen2021evaluating}. We evaluate the functional correctness of generated programs using the benchmark's original execution tests.

\paragraph{LiveCodeBench.}
LiveCodeBench continuously collects new programming contest problems from LeetCode, AtCoder, and Codeforces and provides problem release dates for time-based evaluation\citep{jain2025livecodebench}. We use its code-generation scenario and report the benchmark release and problem release-date window used in our evaluation.

These datasets serve as evaluation benchmarks. Their evaluation problems, reference solutions, and near-duplicate variants are excluded from the distillation training data. We report final-answer accuracy for mathematics and pass@1 on the augmented test suites for code generation.

\subsubsection{Baselines}
We compare our method with the following state-of-the-art approaches.

\paragraph{Trace Inversion.}
Trace Inversion trains an inversion model to reconstruct detailed reasoning from questions, answers, and optional reasoning summaries \citep{zhang2026steal}. The reconstructed reasoning is subsequently used to fine-tune a student model, providing a comparison for acquiring distillation supervision from black-box teacher outputs.

\paragraph{Answer-Conditioned CoT.}
Answer-Conditioned CoT examines how answer conditioning affects reasoning distillation \citep{lee2026answerconditioned}. It compares question-only generation (Blind CoT) with generation conditioned
on a gold answer or a reference implementation (Leaked CoT). Both conditions apply the same correctness filter and use matched training problems and an identical SFT recipe.

\paragraph{GLOW.}
GLOW (Gain-based Loss Weighting) uses a mixture of reasoning trajectories with correct and incorrect final answers and adjusts sample loss weights according to their loss reductions across epochs \citep{xueyun2026glow}. It provides a comparison for the student-training strategy.

Trace Inversion and Answer-Conditioned CoT concern the construction of reasoning supervision, while GLOW modifies the weighting of the student-training loss.

\subsection{Results and Analysis}
\subsubsection{Experimental Results}

Table~\ref{tab:main_results} compares our method with Trace Inversion, Answer-Conditioned CoT, and GLOW. Ours achieves the highest reported score in 14 of the 16 student--teacher--benchmark combinations. With GLM-5.2, Ours ranks first on all four benchmarks for both students. With DeepSeek-V4-Pro, two exceptions occur: GLOW scores higher on LiveCodeBench for Qwen2.5-7B-Instruct (40.5 versus 40.1), and Trace Inversion scores higher on GSM8K for Llama-3.1-8B-Instruct (90.4 versus 90.1).

\begin{table*}[t]
\centering
\caption{Comparison with the three published baselines across two students, two teachers, and four benchmarks. Mathematics scores are accuracy (\%), and code-generation scores are pass@1 (\%). Higher is better. The best score within each student--teacher--benchmark combination is shown in bold.}
\label{tab:main_results}
\small
\setlength{\tabcolsep}{5pt}
\renewcommand{\arraystretch}{1.08}
\resizebox{\textwidth}{!}{%
\begin{tabular}{@{}llrrrr@{}}
\toprule
\multirow{2}{*}{Teacher} & \multirow{2}{*}{Method}
& \multicolumn{2}{c}{Mathematics} & \multicolumn{2}{c}{Code Generation} \\
\cmidrule(lr){3-4}\cmidrule(lr){5-6}
& & MATH-500 & GSM8K & HumanEval & LiveCodeBench \\
\midrule
\multicolumn{6}{c}{\textbf{Student: Qwen2.5-7B-Instruct}} \\
\midrule
\multirow{4}{*}{DeepSeek-V4-Pro} & Trace Inversion & 75.4 & 94.1 & 91.0 & 39.5 \\
 & Answer-Conditioned CoT & 72.3 & 92.0 & 87.9 & 34.2 \\
 & GLOW & 77.6 & 93.7 & 90.7 & \textbf{40.5} \\
 & \textbf{Ours} & \textbf{79.3} & \textbf{94.3} & \textbf{92.2} & 40.1 \\
\cmidrule(lr){1-6}
\multirow{4}{*}{GLM-5.2} & Trace Inversion & 77.4 & 93.1 & 88.3 & 40.5 \\
 & Answer-Conditioned CoT & 74.0 & 90.7 & 82.8 & 35.1 \\
 & GLOW & 78.1 & 92.4 & 87.6 & 39.8 \\
 & \textbf{Ours} & \textbf{78.9} & \textbf{93.8} & \textbf{88.9} & \textbf{41.6} \\
\midrule
\multicolumn{6}{c}{\textbf{Student: Llama-3.1-8B-Instruct}} \\
\midrule
\multirow{4}{*}{DeepSeek-V4-Pro} & Trace Inversion & 57.1 & \textbf{90.4} & 87.3 & 16.7 \\
 & Answer-Conditioned CoT & 53.2 & 86.1 & 81.2 & 13.9 \\
 & GLOW & 58.4 & 89.3 & 87.8 & 17.5 \\
 & \textbf{Ours} & \textbf{59.9} & 90.1 & \textbf{88.4} & \textbf{18.4} \\
\cmidrule(lr){1-6}
\multirow{4}{*}{GLM-5.2} & Trace Inversion & 60.9 & 90.7 & 84.4 & 16.1 \\
 & Answer-Conditioned CoT & 57.0 & 85.5 & 79.3 & 12.9 \\
 & GLOW & 60.1 & 91.1 & 85.5 & 16.7 \\
 & \textbf{Ours} & \textbf{61.1} & \textbf{91.5} & \textbf{85.9} & \textbf{17.2} \\
\bottomrule
\end{tabular}%
}
\end{table*}

Table~\ref{tab:supervision_results} compares Ours with the undistilled checkpoints and the two teacher-supervision settings. Ours obtains a higher reported score than these settings in all 16 student--teacher--benchmark combinations. Relative to the undistilled checkpoints, the absolute gains range from 3.7 to 16.3 percentage points.

\begin{table*}[t]
\centering
\caption{Comparison of supervision settings. Metrics and units follow Table~\ref{tab:main_results}. Undistilled denotes the original student checkpoint and is reported once per student because it is independent of the teacher. Best scores in each teacher-specific comparison, including the undistilled reference, are shown in bold.}
\label{tab:supervision_results}
\small
\setlength{\tabcolsep}{5pt}
\renewcommand{\arraystretch}{1.08}
\resizebox{\textwidth}{!}{%
\begin{tabular}{@{}llrrrr@{}}
\toprule
\multirow{2}{*}{Teacher} & \multirow{2}{*}{Method}
& \multicolumn{2}{c}{Mathematics} & \multicolumn{2}{c}{Code Generation} \\
\cmidrule(lr){3-4}\cmidrule(lr){5-6}
& & MATH-500 & GSM8K & HumanEval & LiveCodeBench \\
\midrule
\multicolumn{6}{c}{\textbf{Student: Qwen2.5-7B-Instruct}} \\
\midrule
-- & Undistilled & 70.9 & 90.1 & 82.1 & 29.1 \\
\cmidrule(lr){1-6}
\multirow{3}{*}{DeepSeek-V4-Pro} & Teacher CoT & 75.3 & 93.9 & 90.7 & 36.9 \\
 & Teacher Answer Explanation & 72.1 & 91.5 & 88.3 & 33.7 \\
 & \textbf{Ours} & \textbf{79.3} & \textbf{94.3} & \textbf{92.2} & \textbf{40.1} \\
\cmidrule(lr){1-6}
\multirow{3}{*}{GLM-5.2} & Teacher CoT & 76.1 & 92.9 & 86.3 & 35.3 \\
 & Teacher Answer Explanation & 73.4 & 90.8 & 83.7 & 34.2 \\
 & \textbf{Ours} & \textbf{78.9} & \textbf{93.8} & \textbf{88.9} & \textbf{41.6} \\
\midrule
\multicolumn{6}{c}{\textbf{Student: Llama-3.1-8B-Instruct}} \\
\midrule
-- & Undistilled & 49.6 & 83.2 & 72.1 & 10.9 \\
\cmidrule(lr){1-6}
\multirow{3}{*}{DeepSeek-V4-Pro} & Teacher CoT & 55.4 & 88.7 & 86.3 & 13.8 \\
 & Teacher Answer Explanation & 53.7 & 85.3 & 79.9 & 13.0 \\
 & \textbf{Ours} & \textbf{59.9} & \textbf{90.1} & \textbf{88.4} & \textbf{18.4} \\
\cmidrule(lr){1-6}
\multirow{3}{*}{GLM-5.2} & Teacher CoT & 57.3 & 89.1 & 82.4 & 12.6 \\
 & Teacher Answer Explanation & 56.7 & 84.9 & 78.6 & 11.4 \\
 & \textbf{Ours} & \textbf{61.1} & \textbf{91.5} & \textbf{85.9} & \textbf{17.2} \\
\bottomrule
\end{tabular}%
}
\end{table*}

During our experiments, we observed that teacher-generated reasoning trajectories can contain errors. In our evaluated settings, students fine-tuned on unfiltered distillation data outperformed those trained only on trajectories with correct final answers. This observation is consistent with prior evidence that final-answer correctness alone does not fully determine the usefulness of reasoning supervision. \citet{xueyun2026glow} show that incorporating trajectories labeled as incorrect can improve out-of-domain generalization over
positive-only supervised fine-tuning, as such trajectories may still contain valid intermediate reasoning. \citet{li2025structure} find that demonstrations with incorrect answers remain effective for learning long-chain-of-thought reasoning, whereas shuffling or deleting steps causes greater performance degradation. Similarly, \citet{lippmann2025style} report improvements over base models when synthetic reasoning traces preserve stylistic patterns but lead to incorrect final answers, although their correct-answer counterparts
perform better. Together, these findings suggest that traces with incorrect final answers can still provide useful intermediate supervision and preserve reasoning structure. To examine this local supervisory value, we introduce \textbf{Stepwise Explanation Information Gain (SEIG)}, which measures the log-ratio of
a frozen student's estimated probabilities of generating a correct step conclusion under a stepwise explanation and the corresponding explanation from a globally generated CoT. Both conditions use the same problem, verified premises, and step task, with direct leakage of the target conclusion removed. A positive SEIG indicates that the stepwise explanation is more helpful for correct step completion, which may hold even when the full trajectory ends with an incorrect answer. SEIG therefore provides a diagnostic for useful local
supervision within imperfect trajectories; its relationship to downstream fine-tuning gains requires separate empirical validation.

\subsubsection{Diagnostic Analysis with SEIG}

\begin{figure*}[!tp]
    \centering
    \includegraphics[width=\textwidth]{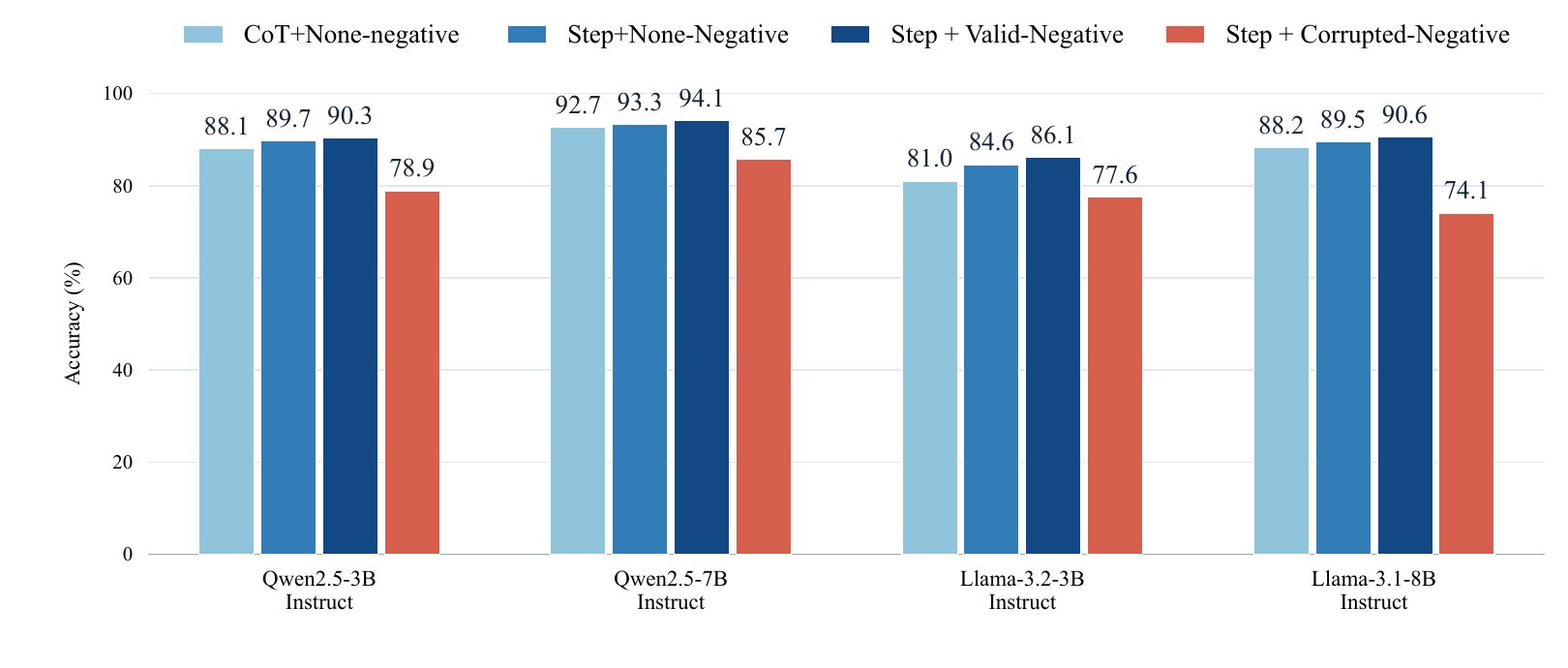}
    \caption{
    GSM8K accuracy (\%) after SFT under four training-data configurations. COT+None-negative and Step+None-Negative use conventional CoT and our stepwise trajectories, respectively, retaining only those with correct final answers. Step + Valid-Negative and Step + Corrupted-Negative mix correct-answer stepwise trajectories with incorrect-answer trajectories containing many and few valid intermediate steps, respectively.
    }
    \label{fig:seig1}
\end{figure*}

\paragraph{Value of Incorrect-Answer Trajectories.}
To assess the rationale behind SEIG, we fine-tune four student models using the training-data configurations defined in Figure~\ref{fig:seig1} and evaluate them on GSM8K. \emph{Step + Valid-Negative} achieves the highest accuracy across all four students, outperforming \emph{Step+None-Negative} by 0.6--1.5 percentage points. \emph{Step + Corrupted-Negative}, however, reduces accuracy relative to \emph{Step+None-Negative} for every student. 
These results suggest that trajectories with incorrect final answers can still provide useful supervision when they preserve substantial valid intermediate reasoning, supporting SEIG's focus on local supervisory value beyond final-answer correctness.

\begin{figure*}[!tp]
    \centering
    \includegraphics[width=\textwidth]{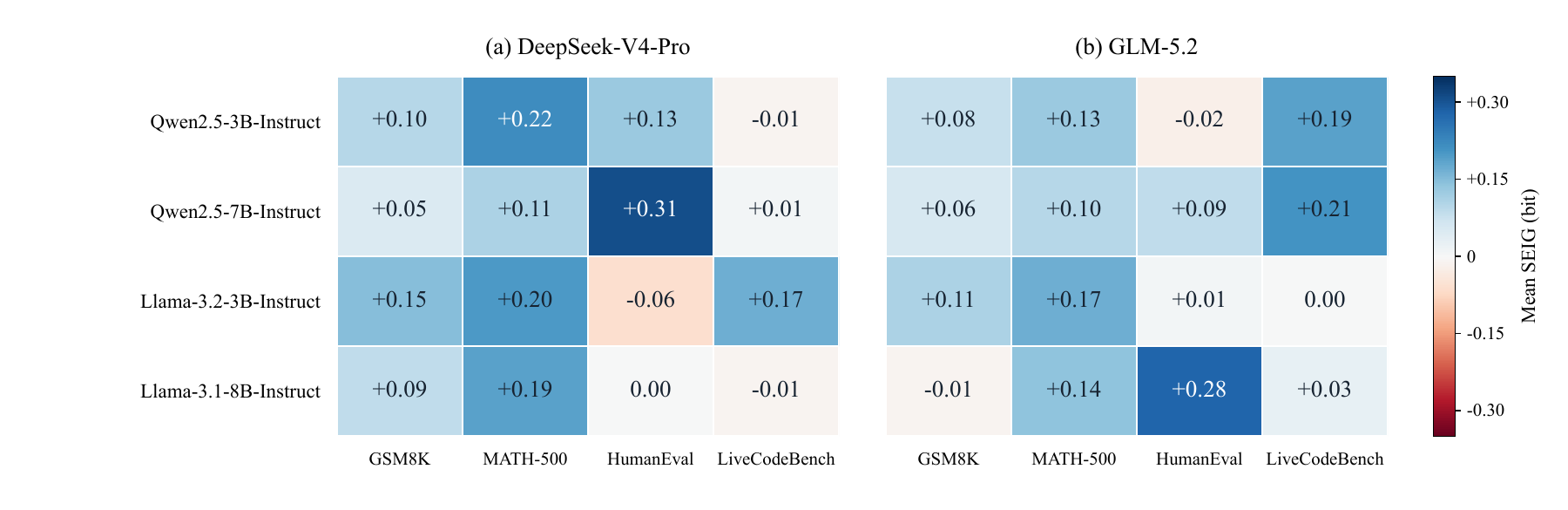}
    \caption{
    Mean SEIG (bits) of stepwise explanations relative to global CoT explanations. Panels (a) and (b) correspond to DeepSeek-V4-Pro and GLM-5.2, respectively; rows represent student checkpoints and columns represent datasets. For each shared step, CoT and Step each sample $K=16$ teacher-generated reasoning trajectories, whose local utility is evaluated using the same frozen student checkpoint and common local conditions. Positive values (blue) favor stepwise explanations, whereas negative values (red) favor global CoT explanations.
    }
    \label{fig:seig2}
\end{figure*}

\paragraph{SEIG across Teachers, Students, and Tasks.}
With $K=16$ teacher-generated reasoning trajectories per step for each of CoT and Step, Figure~\ref{fig:seig2} shows positive mean SEIG in 25 of the 32 evaluated teacher--student--dataset combinations. Although five combinations exhibit negative means, their absolute magnitudes are modest, ranging from 0.01 to 0.06, whereas positive gains reach 0.31 bits. Gains are particularly consistent on MATH-500, where all eight combinations yield positive values ranging from 0.10 to 0.22 bits. Overall, this predominantly positive pattern supports the advantage of our stepwise method for local step completion across the evaluated teachers, students, and tasks, despite modest negative values in a limited number of settings.

\begin{figure*}[!tp]
    \centering
    \includegraphics[width=\textwidth]{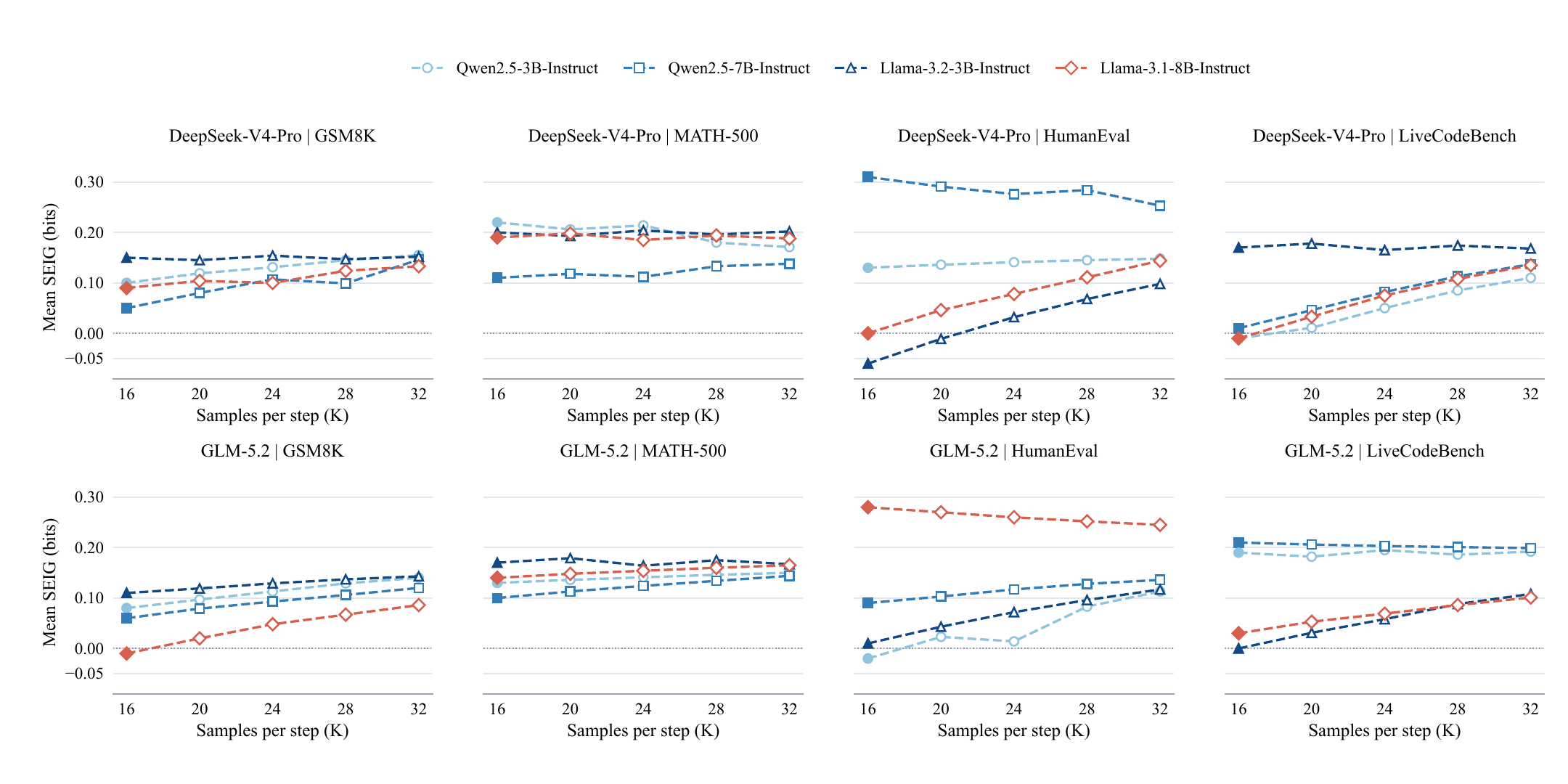}
    \caption{
    Mean SEIG (bits) across teacher sampling budgets $K\in\{16,20,24,28,32\}$. Rows correspond to DeepSeek-V4-Pro and GLM-5.2, columns represent datasets, and curves represent student checkpoints. For each step, both CoT and Step sample $K$ teacher-generated reasoning trajectories, whose local utility is evaluated using
    the same frozen student checkpoint and common local conditions. SEIG compares the two methods under the same teacher sampling budget.
    }
    \label{fig:seig3}
\end{figure*}

\paragraph{Sensitivity to the Teacher Sampling Budget.}
Figure~\ref{fig:seig3} examines mean SEIG across teacher sampling budgets $K\in\{16,20,24,28,32\}$. Positive mean SEIG is observed in 25 of the 32 teacher--student--dataset combinations at $K=16$ and in all
32 combinations for $K\geq24$. As $K$ increases, initially negative estimates become positive, while some large initial estimates decrease and other curves exhibit modest fluctuations. These variations show that increasing the teacher sampling budget does not uniformly increase SEIG.

\begin{figure*}[!tp]
    \centering
    \includegraphics[width=\textwidth]{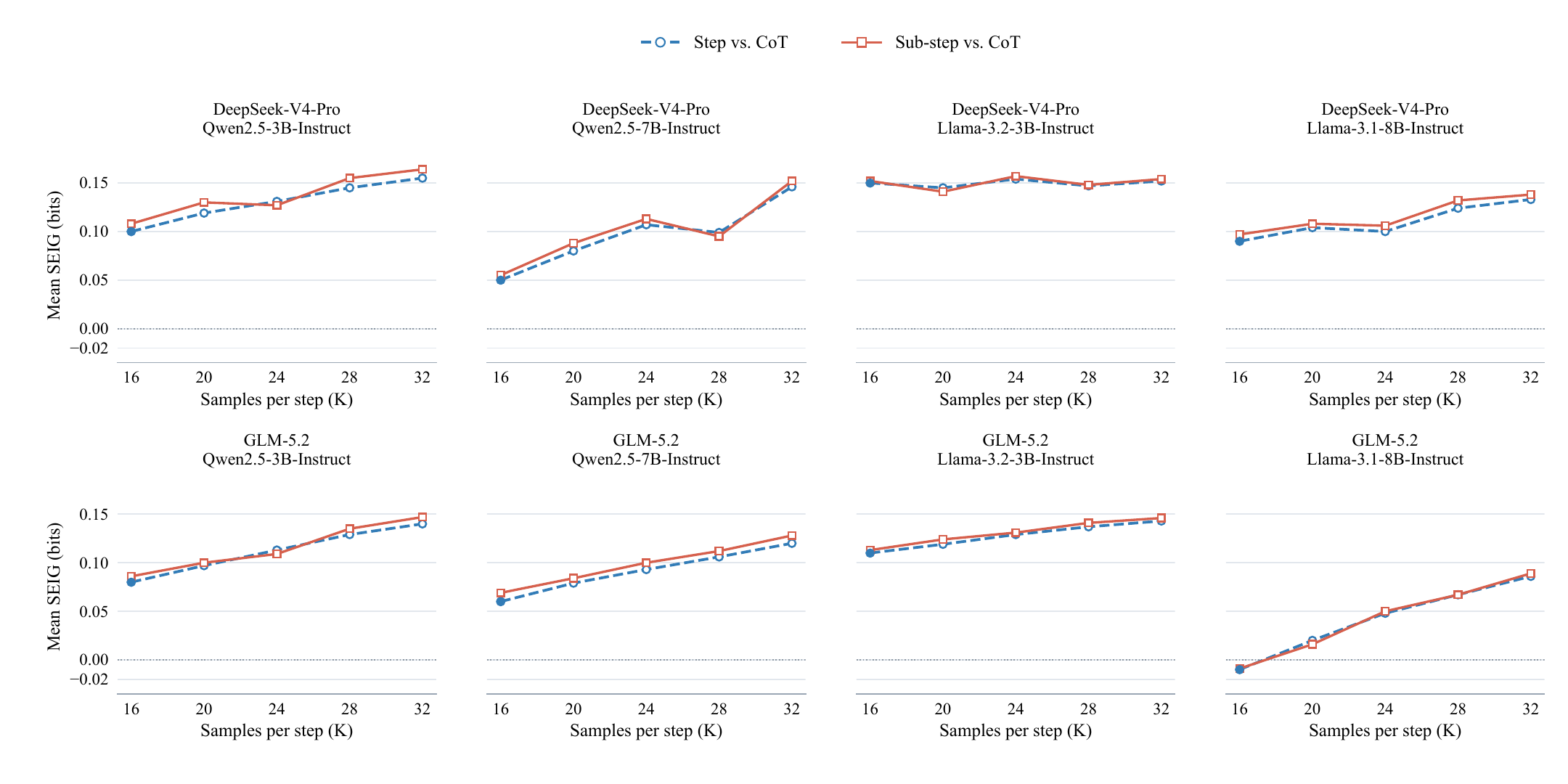}
    \caption{
    Mean SEIG (bits) of Step and Sub-step relative to the same CoT baseline on GSM8K. Sub-step applies one additional round of decomposition to Step explanations. Rows correspond to teachers and columns to student checkpoints. For each evaluated parent step, each configuration samples $K\in\{16,20,24,28,32\}$ teacher-generated reasoning trajectories; comparisons use the same frozen student checkpoint and common local evaluation conditions.
    }
    \label{fig:seig4}
\end{figure*}

\begin{figure*}[!tp]
    \centering
    \includegraphics[width=\textwidth]{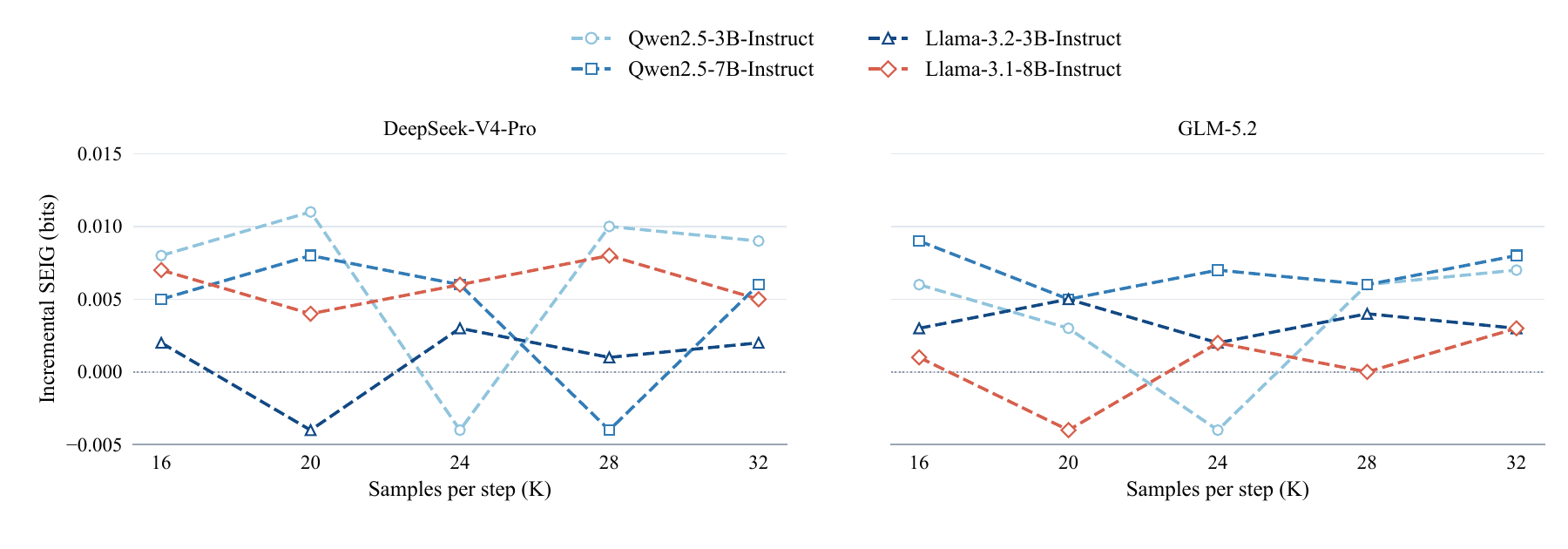}
    \caption{
    Incremental SEIG (bits) from further decomposition on GSM8K, computed as the mean SEIG of Sub-step versus CoT minus that of Step versus CoT under the same evaluation conditions and teacher sampling budget $K$.
    Panels correspond to teachers and curves to student checkpoints. The dotted horizontal line marks zero incremental gain; positive and negative values indicate that Sub-step yields higher and lower mean SEIG than Step, respectively.
    }
    \label{fig:seig4_incremental}
\end{figure*}

\paragraph{Diminishing Returns from Recursive Decomposition.}
Having established the utility of Step, we investigate whether an additional round of decomposition into Sub-step explanations provides further information gain. On GSM8K, Figure~\ref{fig:seig4} shows that Sub-step closely tracks Step relative to the same CoT baseline across both teachers, all four students, and the tested teacher sampling budgets. Figure~\ref{fig:seig4_incremental} further shows that the incremental SEIG remains small, with occasional slightly negative values and no consistent increase with $K$. These observations suggest diminishing returns from recursive decomposition, with most of the gains already captured at the
Step level in these settings.

One possible explanation is that further refinement is constrained by the evaluated teachers' ability to generate additional useful intermediate reasoning: finer decomposition may introduce redundant detail, yielding little additional information for the student.

\section{Conclusion}
We presented Stepwise-Querying, a distillation framework that constructs detailed reasoning supervision from a black-box teacher without access to its internal chain-of-thought traces. Starting from a brief candidate solution, our protocol issues independent, step-specific queries whose prompts ask the teacher to derive each step from the problem, check validity and completeness, and propose corrections, treating the initial answer as provisional rather than as a premise. The assembled solutions are then used for standard supervised fine-tuning. 
Across two teachers, two students, and four mathematics and code benchmarks, our method achieves the highest score in 14 of 16 comparisons against three published baselines and consistently outperforms global CoT and answer-explanation supervision under identical training recipes. We further introduced Stepwise Explanation Information Gain (SEIG), a diagnostic that measures, under a frozen student and shared local conditions, the log-ratio of correct step-completion probabilities between stepwise and global CoT explanations, requiring only public explanation text and independent verification labels. SEIG analyses yield two findings: trajectories with incorrect final answers remain useful when they preserve valid intermediate reasoning, and recursive decomposition beyond the step level yields diminishing returns. Together, these results suggest that the locality and validity of intermediate supervision—rather than final-answer correctness alone—is a key driver of distillation quality. Future work includes validating the relationship between SEIG and downstream fine-tuning gains, extending the protocol to logical inference and multi-hop question answering, and reducing the number of teacher calls required per problem, for example by adaptively selecting which steps to expand.






\bibliography{custom}

\appendix

\section{Appendix}
\label{sec:appendix}

\begin{figure*}[!tp]
    \centering
    \includegraphics[width=0.9\textwidth]{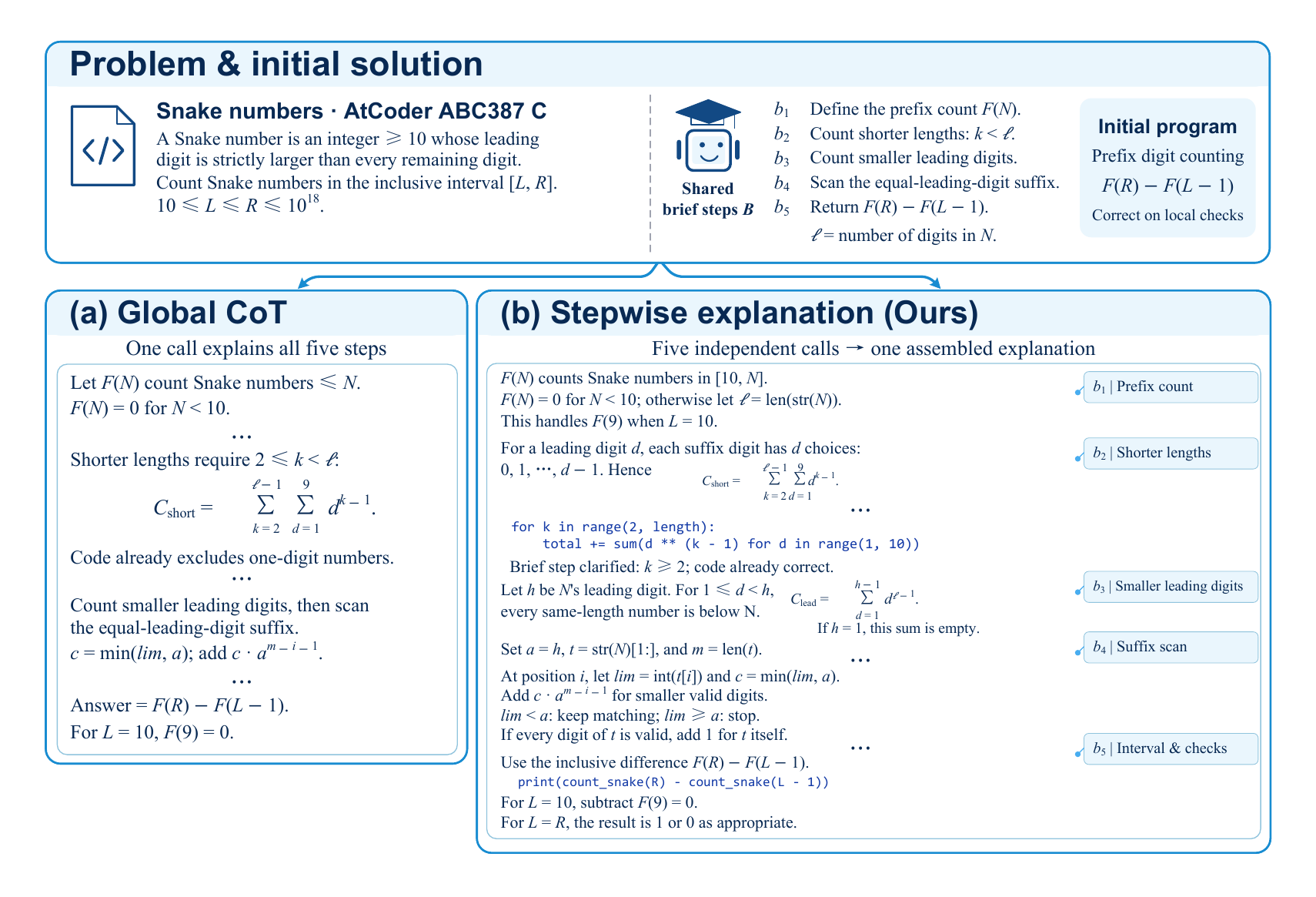}
    \caption{The figure compares the model’s chain of thought with the stepwise explanations generated by our method for a mathematical problem.}
    \label{fig:apx1}
\end{figure*}

\begin{figure*}[!tp]
    \centering
    \includegraphics[width=0.9\textwidth]{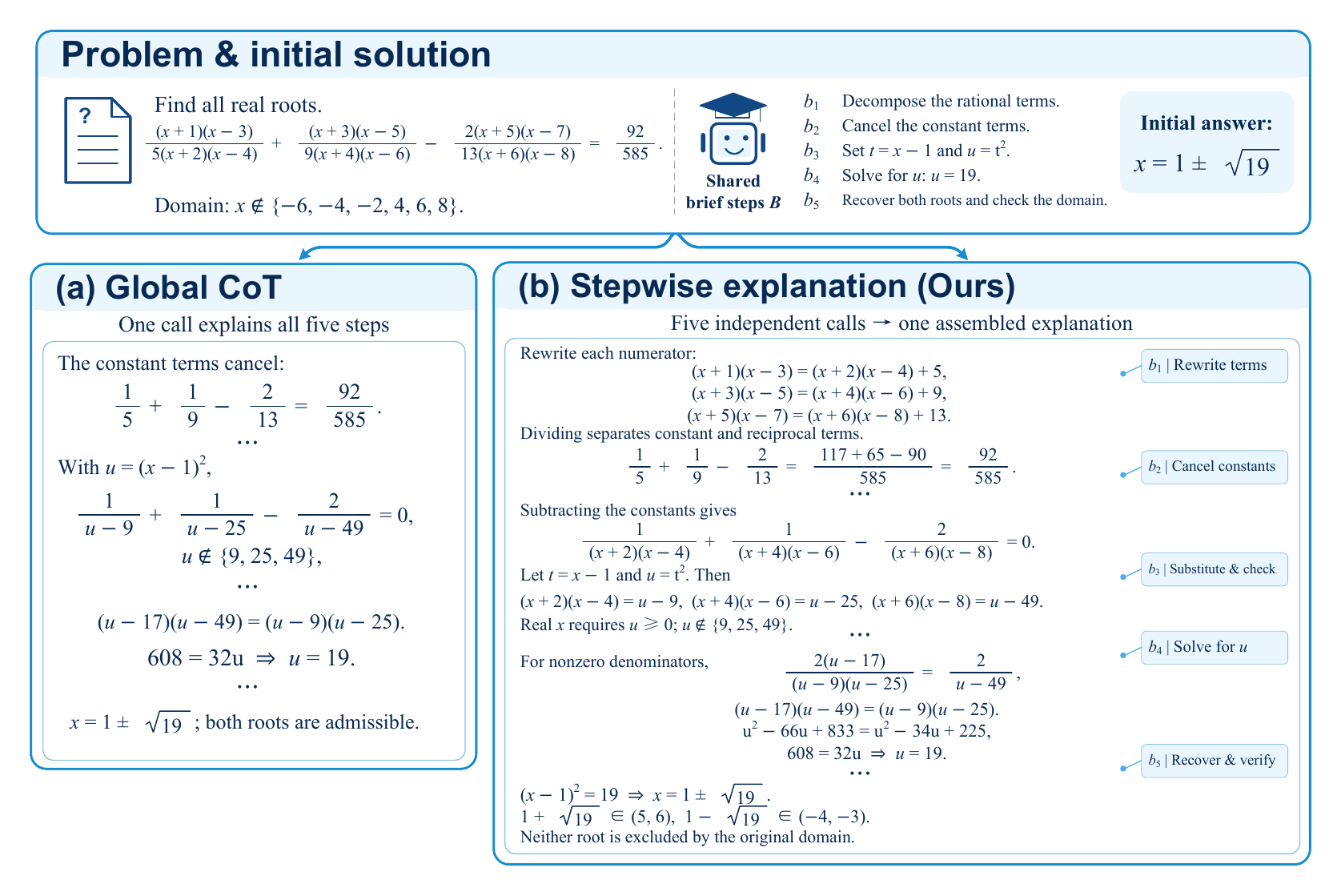}
    \caption{The figure compares the model’s chain of thought with the stepwise explanations generated by our method for a code problem.}
    \label{fig:apx2}
\end{figure*}

We select one example each from MATH-500 and LiveCodeBench to illustrate our method in Figures~\ref{fig:apx1} and ~\ref{fig:apx2}, respectively. For clarity, only the key steps are shown. These examples suggest that our method provides more informative explanations at critical steps without relying on the model’s internal chain-of-thought traces.

\end{document}